\documentclass[runningheads]{llncs}

\usepackage{amsmath,amssymb} 

\usepackage{color}

\usepackage[pagebackref=true,breaklinks=true,letterpaper=true,colorlinks,bookmarks=false]{hyperref}

\usepackage{graphics}
\usepackage{subfigure}
\usepackage{epsfig}
\usepackage{epstopdf}

\usepackage{algorithm}
\usepackage[noend]{algpseudocode}

\usepackage{graphicx}
\usepackage{amsmath,amssymb} 
\usepackage{color}

\DeclareMathOperator*{\argmin}{arg\,min}
\begin{document}

\newcommand{\point}{
    \raise0.7ex\hbox{.}
    }
\pagestyle{headings}
\mainmatter


\title{Sparse Coding on Cascaded Residuals}



\title{Sparse Coding on Cascaded Residuals} 

\titlerunning{Sparse Coding on Cascaded Residuals} 

\authorrunning{Tong Zhang, Fatih Porikli} 

\author{Tong Zhang$^1$, Fatih Porikli$^{1,2}$} 
\institute{Australian National University$^1$, Data61/CSIRO$^2$ \\
\{\textbf{tong.zhang, fatih.porikli\}@anu.edu.au}} 

\maketitle

\begin{abstract}
This paper seeks to combine dictionary learning and hierarchical image representation in a principled way. To make dictionary atoms capturing additional information from extended receptive fields and attain improved descriptive capacity, we present a two-pass multi-resolution cascade framework for dictionary learning and sparse coding. The cascade allows collaborative reconstructions at different resolutions using the same dimensional dictionary atoms. Our jointly learned dictionary comprises atoms that adapt to the information available at the coarsest layer where the support of atoms reaches their maximum range and the residual images where the supplementary details progressively refine the reconstruction objective. The residual at a layer is computed by the difference between the aggregated reconstructions of the previous layers and the downsampled original image at that layer. Our method generates more flexible and accurate representations using much less number of coefficients. Its computational efficiency stems from encoding at the coarsest resolution, which is minuscule, and encoding the residuals, which are relatively much sparse. Our extensive experiments on multiple datasets demonstrate that this new method is powerful in image coding, denoising, inpainting and artifact removal tasks outperforming the state-of-the-art techniques. 
\end{abstract}


\section{Introduction}

Sparse representation promises noise resilience by assigning representation coefficients from dictionary atoms characterizing the clean data distribution, improved classification performance by attaining discriminative features, robustness by preventing the model from overfitting data, and semantic interpretation by allowing atoms to associate with meaningful attributes. Computer vision applications include image compression, regularization in reverse problems, feature extraction, recognition, interpolation for incomplete data, and more~\cite{mairal2009non,Aharon2006,mairal2009supervised,yan2013nonlocal,Ophir2011,sulam2014image}.

An overcomplete dictionary that leads to sparse representations can either be chosen from a predetermined set of functions or designed by adapting its content to fit a given set of samples. The performance of the predetermined dictionaries, such as overcomplete Discrete Cosine Transform (DCT)~\cite{ahmed1974discrete}, wavelets~\cite{mallat1999wavelet}, curvelets~\cite{candes2000curvelets}, contourlets~\cite{do2005contourlet}, shearlets~\cite{labate2005sparse} and other analytic forms, depends on how suitable they are to sparsely describe the samples in question. On the other hand, the learned dictionaries are data driven and tailored for distinct applications. Noteworthy algorithms of this type include the Method of Optimal Directions (MOD)~\cite{Engan1999}, generalized PCA~\cite{Sastry2003}, KSVD~\cite{Aharon2006}, Online Dictionary Learning (ODL)~\cite{Mairal2009,yan2013nonlocal}. The learned dictionaries adapt better compared to analytic ones and provide improved performance.

In general, image based dictionary learning and sparse encoding tasks are formulated as an optimization problem
\begin{align}\label{cost1}
\begin{split}
& \argmin_{\mathbf{D},\mathbf{X}} \| \mathbf{Y}- \mathbf{D}\mathbf{X}  \|_F^2 \;\;\;\;\;\;\;\;
\text{s.t.}  \|\mathbf{x}_i\|_0 \leq T \;,
\end{split}
\end{align}
or its equivalent form
\begin{align}\label{cost2}
\begin{split}
& \argmin_{\mathbf{X},\mathbf{D}} \sum_i \| \mathbf{x}_i \|_0  \;\;\;\;\;\;\;\;
\text{s.t.} \| \mathbf{Y}- \mathbf{D}\mathbf{X}  \|_F^2  \leq \epsilon
\end{split}
\end{align}
where $\mathbf{Y} \in \mathbb{R}^{n \times k}$ is $k$ image patches with dimension $n$, $\mathbf{X} \in \mathbb{R}^{m \times k}$ denotes the coefficients of corresponding images, $\mathbf{D} \in \mathbb{R}^{n \times m}$ is an overcomplete matrix, $T$ is the number of coefficient used to describe the images, and $\epsilon$ is the error tolerance such that once the reconstruction error is smaller than the tolerance the pursuit will be terminated. The sparsity is achieved because $n \ll m$ and $T \ll m$. For an extended discussion on the solutions of above objectives, see Section~\ref{section:relatedwork}.

For mathematical convenience, dictionary learning methods often employ in uniform spaces, e.g. in the vector space of 8$\times$8 image patches. In other words, same scale blocks are pulled from overlapping or non-overlapping image patches on a dense grid and a single-scale dictionary is learned. However, dictionary atoms learned in this fashion tend to be myopic and blind to global context since such fixed-scale patches only contain local information within their small support. Simply increasing the patch size results in adverse outcomes, i.e. decreased the flexibility of the dictionary to fit data and increased computational complexity. Moreover, optimal patch size varies depending on the underlying texture information. For example, finer partitioning by smaller blocks is preferable for textured regions, yet larger blocks would suit better for smooth areas. Suppose the image to be encoded is a 256$\times$256 flat (e.g. all pixels have the same value) image. Using the conventional 8$\times$8 overlapping blocks would require more than 60K coefficients, yet the same image can be represented using a small number of coefficients of larger patches, even only a single coefficient in the ideal case of the patch has the size of the image. 

As an alternative, multi-scale methods aim to learn dictionaries at different image resolutions for the same patch size using shearlets, wavelets, and Laplacian pyramid~\cite{yan2013nonlocal,Ophir2011,tarquino2014multiscale,liu2015general,yin2015sparse}. A major drawback of these methods is that each layer in the pyramid is either processed independently or in small frequency bands; thus reconstruction errors of coarser layers are projected directly on the finest layer. Such errors cannot be compensated by other layers. This implies, to attain a satisfactory quality, all layers need to be constructed accurately. Instead of learning in different image resolutions, \cite{mairal2008learning} first builds separate dictionaries for quadtree partitioned patches and then zero-pad smaller patches to the largest scale. However, the size of the dictionary learned in this fashion is proportional to the maximum patch size, which prohibits its applicability due to heavy computational load and inflated memory requirements. 

Existing multi-scale dictionary learning methods overlook the redundancy between the layers. As a consequence, larger dictionaries are required, and a high number of coefficients are spent unnecessarily on smooth areas. To the best of our knowledge, no method offers a systematic solution where encodings of the coarser scales progressively enhance the reconstructions of the finer layers.  

\begin{figure*}[t]
  \centering
  \subfigure[original]{
    \includegraphics[width=1.1in]{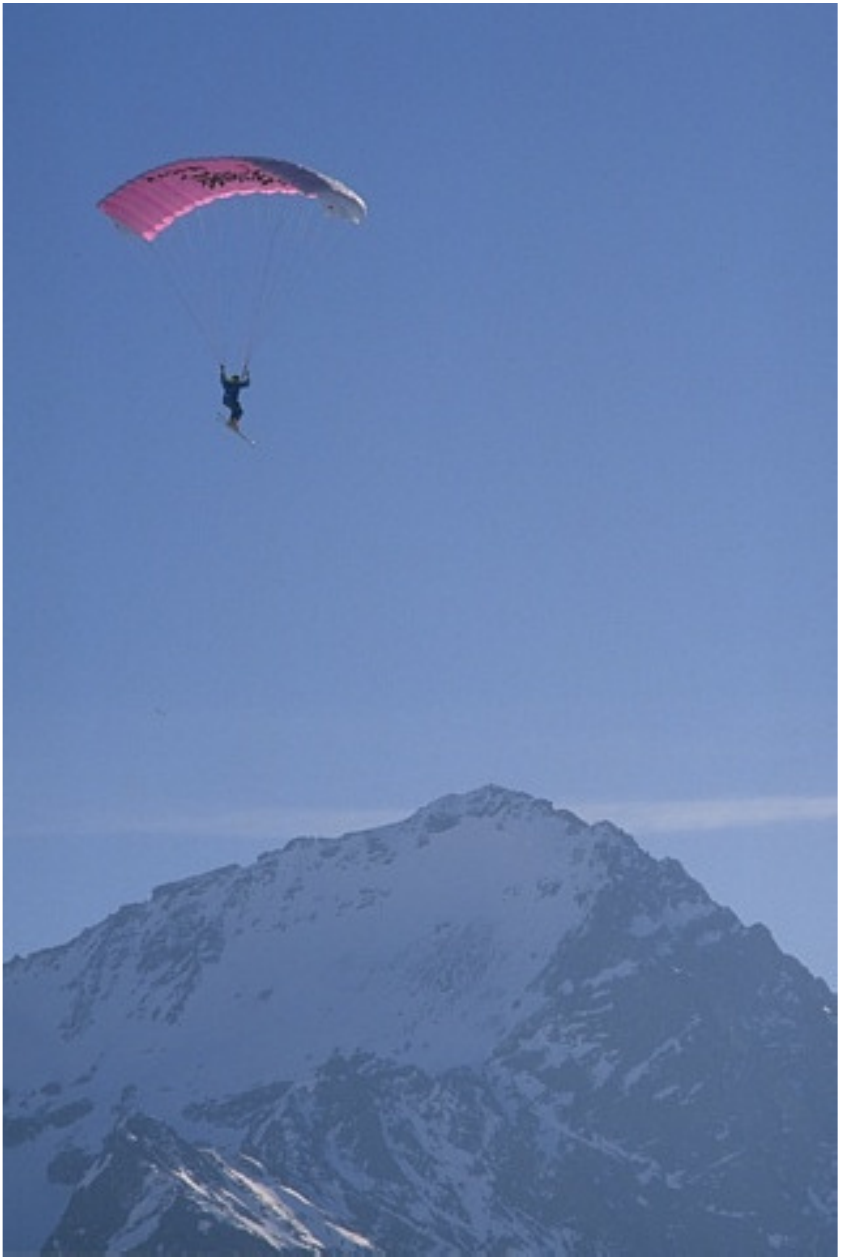}}
  \subfigure[corrupted ]{
    \includegraphics[width=1.1in]{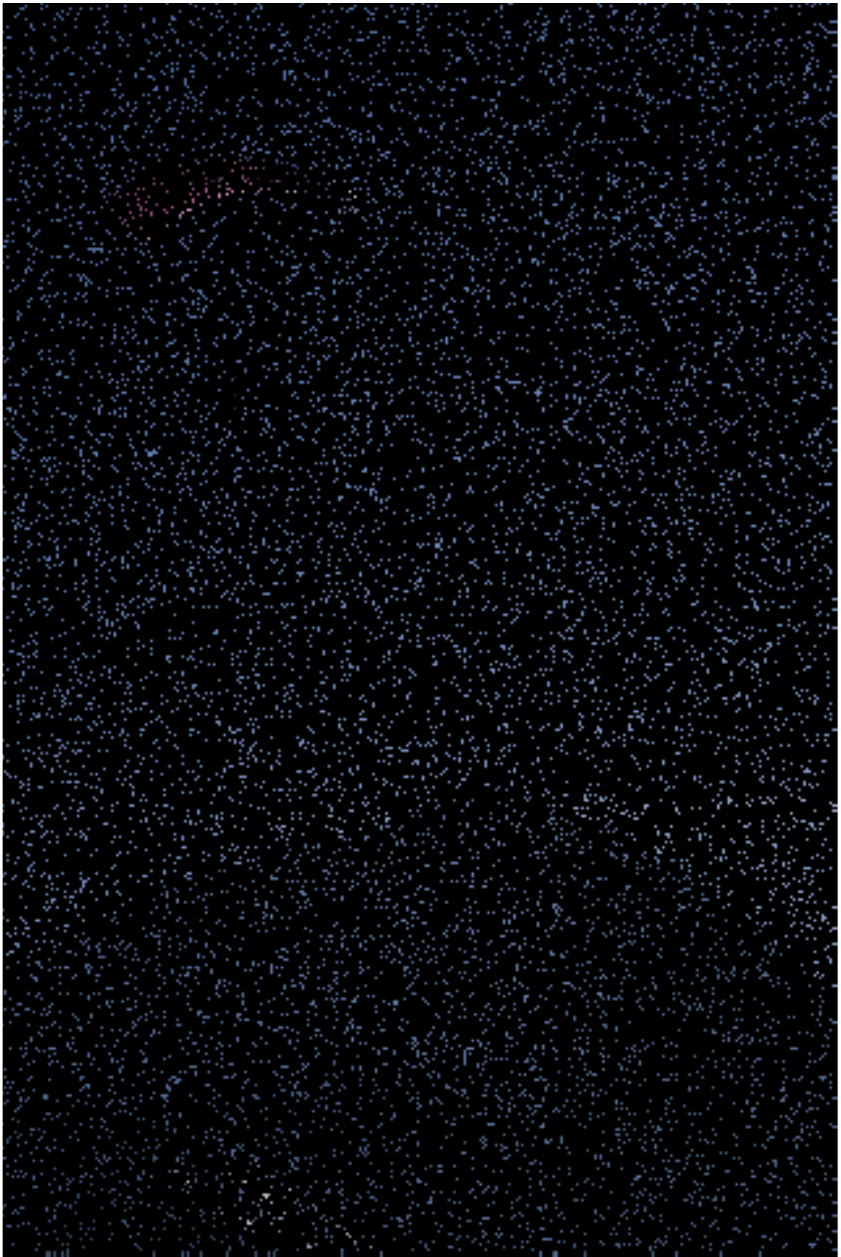}}
  \subfigure[KSVD]{
    \includegraphics[width=1.1in]{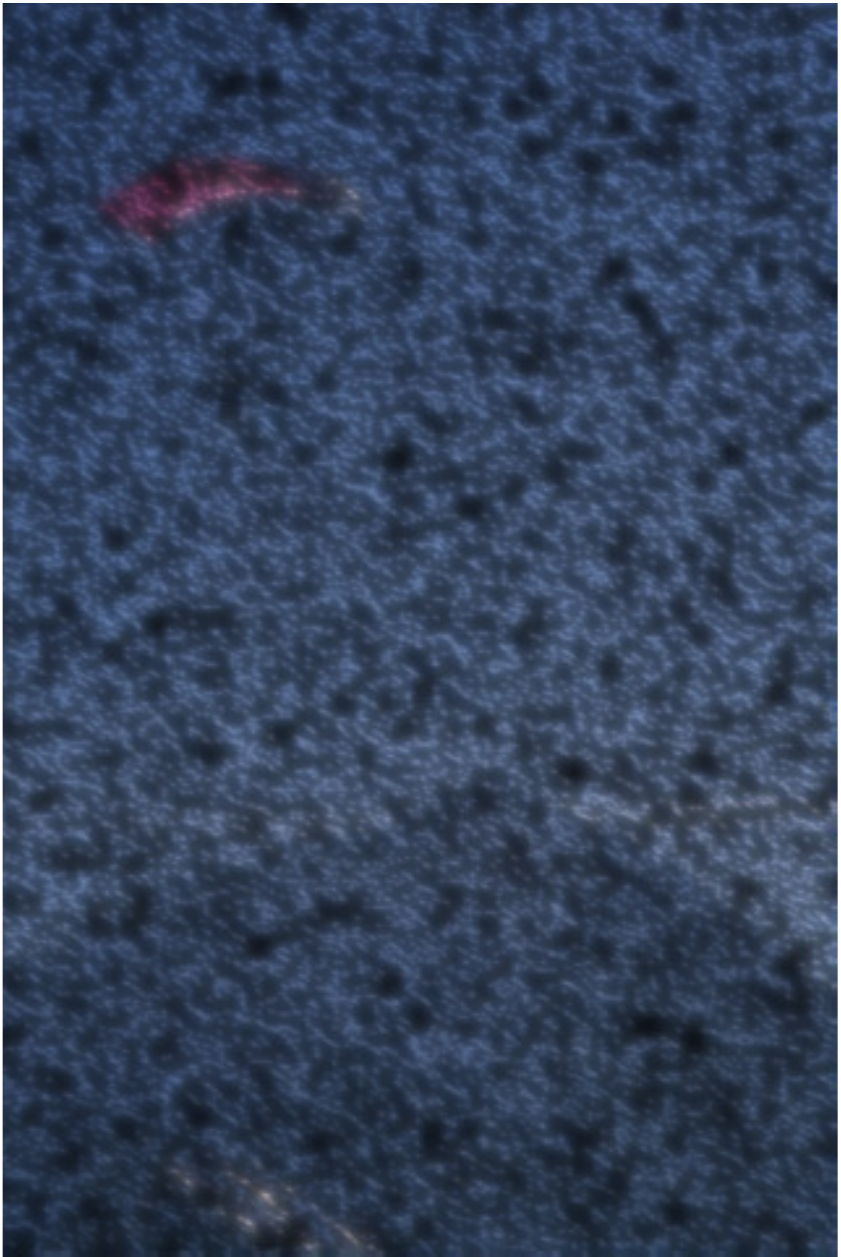}}
    \subfigure[Ours]{
    \includegraphics[width=1.1in]{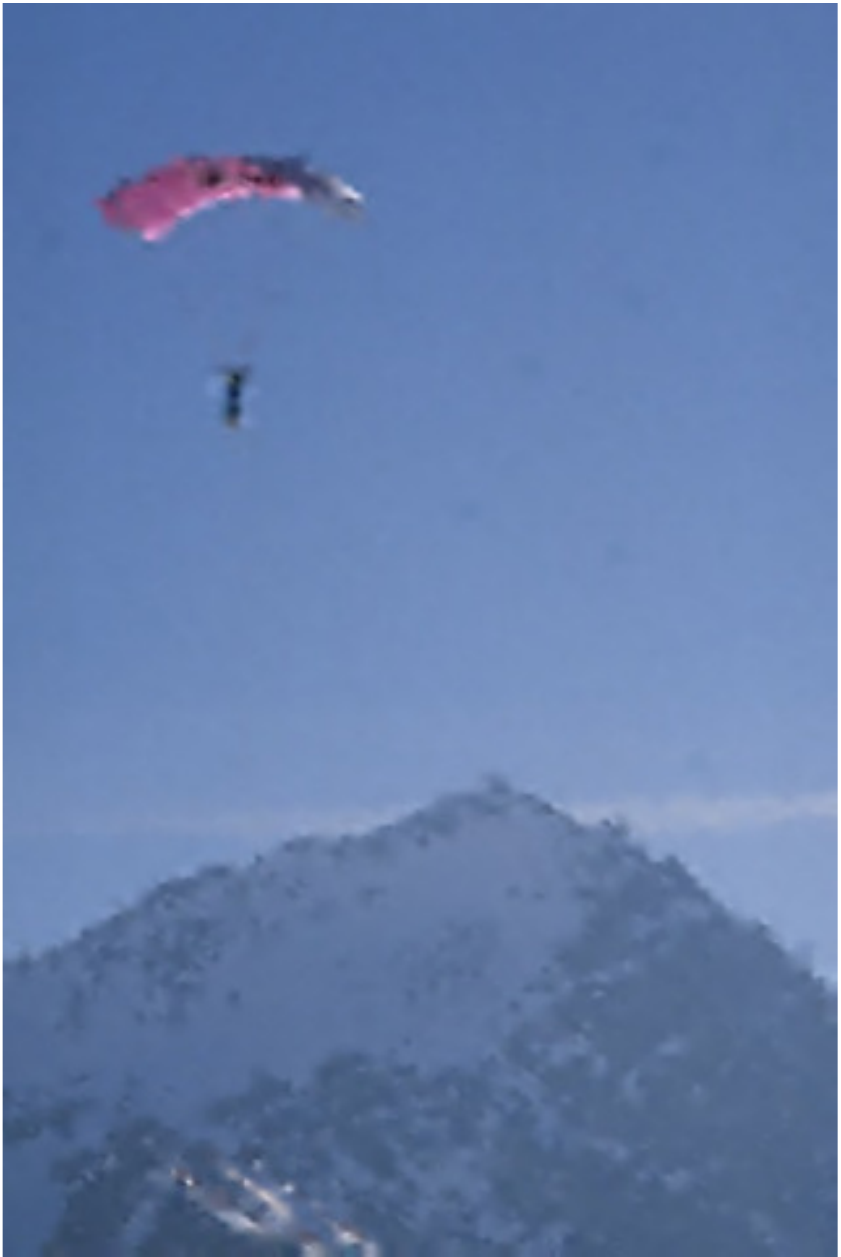}}
    \subfigure[comparsion]{
    \includegraphics[width=4.0in]{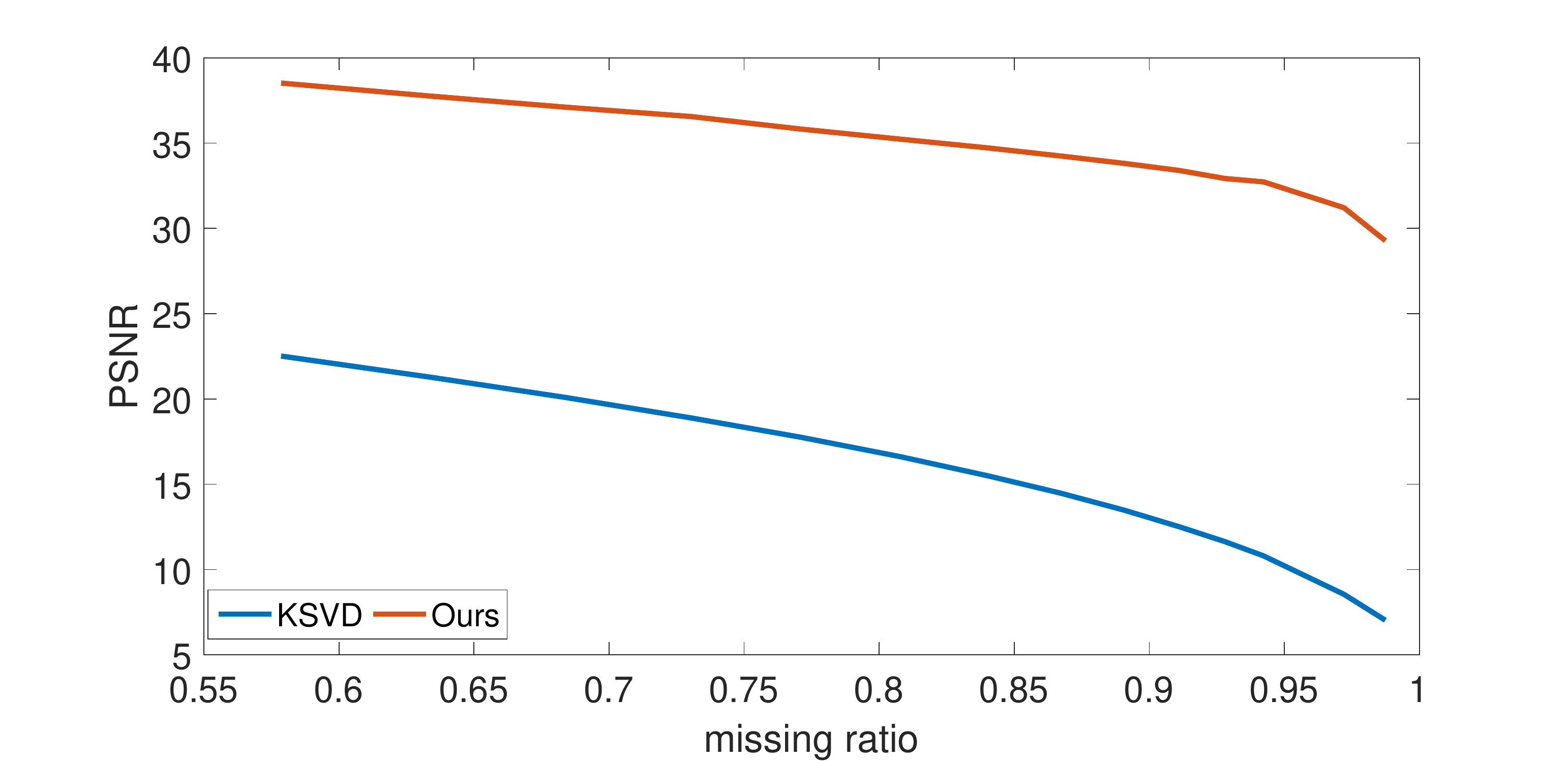}}
  \caption{(a) Original image. (b) Corrupt image where 93$\%$ the original pixels are removed. (c) Reconstruction result of KSVD, PSNR is 11.80 dB. (d) Reconstruction of our method, PSNR is 33.34 dB. (e) Reconstructed image quality vs. the rate of missing coefficients. Red: our method, blue: KSVD. As visible, our method is significantly superior to KSVD.}
  \label{fig:subfig:inpainting2} 
\end{figure*} 

\textbf{Our Contributions}

Aiming to address the above shortcomings and allow dictionary atoms to access larger context for improved descriptive capacity, here we present a computationally efficient cascade framework that employs multi-resolution residual maps for dictionary learning and sparse coding. 

To this end, we start with building an image pyramid using bicubic interpolation. In the first-pass, we learn a dictionary from the coarsest resolution layer and obtain the sparse representation. We upsample the reconstructed image and compute the residual in the next layer. The residual at a level is computed by the difference between the aggregated reconstructions from the coarser layers in a cascade fashion and the downsampled original image at that layer. Dictionaries are learned from the residual in every layer. We use the same patch size yet different resolution input images, which is instrumental in reducing computations and capturing larger context through. The computational efficiency stems from encoding at the coarsest resolution, which is tiny, and encoding the residuals, which are relatively much sparse. This enables our cascade to go as deep as needed without any compromise. 

In the second-pass, we collect all patches from all cascade layers and learn a single dictionary for a final encoding. This naturally solves the problem of determining how many atoms to be assigned at a layer. Thus, the atoms in the dictionary have the same dimension still their receptive fields vary depending on the layer. 

Compared to existing multi-scale approaches operating indiscriminately on image pyramids or wavelets, our dictionary comprises atoms that adapt to the information available at each layer. The supplementary details progressively refine our reconstruction objective. This allows our method to generate a flexible image representation using much less number of coefficients. 

Our extensive experiments on large datasets demonstrate that this new method is powerful in image coding, denoising, inpainting and artifact removal tasks outperforming the state-of-the-art techniques. Figure~\ref{fig:subfig:inpainting2} shows a sample inpainting result from our method where $93\%$ of pixels are missing. As visible, by taking the take advantage of the multi-resolution cascade, we can recover even the very large missing areas.


\section{Sparse Coding on Cascade Layers}
\label{section3}

\begin{figure}
\centering
\scalebox{1.1}[1.1]{\includegraphics[scale = 0.3]{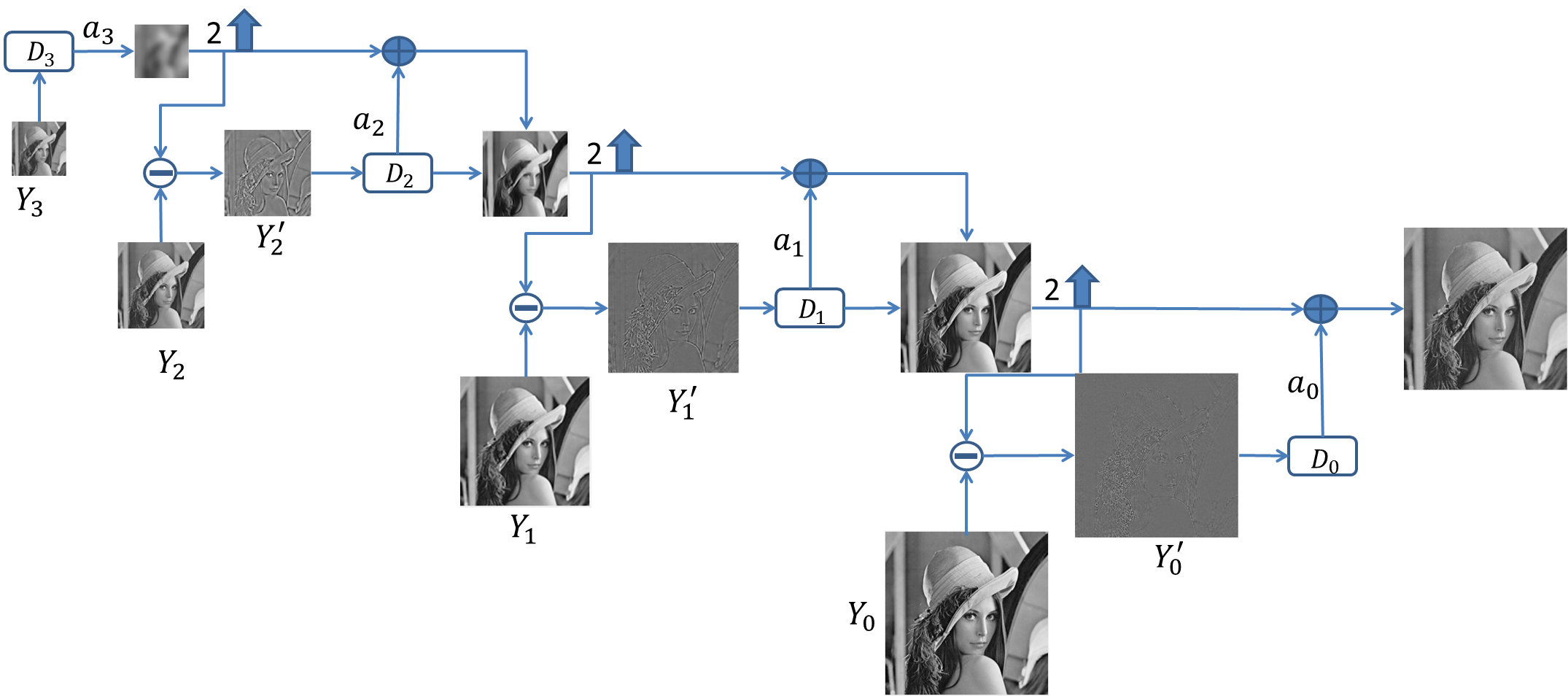}}
\caption{First-pass of our method for a 4-layer cascade. $\mathbf{Y}_0$ is the original image, $\{\mathbf{Y}_3,...,\mathbf{Y}_0 \}$ denote each layer of the image $\mathbf{Y}_3$ pyramid, and $\{\mathbf{D}_3,...,\mathbf{D}_0 \}$ are the  dictionaries. $\mathbf{D}_3$ is learned from the downsampled image, remainig dictionaries are learned from the residuals $\{\mathbf{Y}^{'}_2,\mathbf{Y}^{'}_1,\mathbf{Y}^{'}_0 \}$.
$\mathbf{\alpha}_n$ are the coefficients used to reconstruct each layer.}
\label{structure}
\end{figure}

A flow diagram of our framework is shown in Fig.~\ref{structure} for a sample 4-layer cascade. Given an image $\mathbf{Y}$, we first construct an image pyramid $\mathbf{Y}=$ $\{\mathbf{Y}_0,\mathbf{Y}_1,...\mathbf{Y}_N\}$ by bicubic downsampling. Here, $\mathbf{Y}_0$ is the finest (original) resolution and $\mathbf{Y}_N$ is the coarsest resolution. Other options for the image pyramid are Gaussian pyramid, Laplacian pyramid, bilinear interpolation, and subsampling. Images resampled with bicubic interpolation are smoother and have fewer interpolation artifacts. In contrast, bicubic interpolation considers larger support. 

We employ a two-pass scheme wherein the first-pass we obtain residuals from layer-wise dictionaries, and in the second-pass, we learn a single, global dictionary that extracts and refines the atoms from the dictionaries generated in the first-pass.  

\begin{algorithm}[t]
\caption{Cascade sparse coding}\label{alg:psedo1}
\begin{algorithmic}[1]
\Require\\
{$N$ (the highest pyramid layer), $\mathbf{Y}$(image), \\
$T_n$ (number of coefficient used in layer n)}
\Ensure $\mathbf{Y}^{'}, \hat{\mathbf{Y}},\hat{\mathbf{D}}_{global}$ 
\State $ \mathbf{Y}_n \gets  \text{subsampling}(\mathbf{Y},2^{n})$
\For {$n = \{N,N-1,\cdots,0 \}$}  
\If{n = N}
\State $\mathbf{Y}^{'}_n  \gets \mathbf{Y}_n $
\Else
\State $\mathbf{Y}^{'}_n  \gets \mathbf{Y}_n - \text{upsample}(\hat{\mathbf{Y}}_{n+1},2)$
\EndIf
\State  
\text{Perform KSVD to learn dictionary} $\hat{\mathbf{D}}_n$ and encode  $\mathbf{Y}^{'}_n$
\State $\forall {ij} \
  \{ \hat{\mathbf{x}}^{ij}_n, \hat{\mathbf{D}}_n\} \gets  \argmin\limits_{\mathbf{x}_n^{ij}, \mathbf{D}_n} \sum_{ij}\|\mathbf{R}_{ij}\mathbf{Y}_n^{'}-\mathbf{D}_n\mathbf{x}_n^{ij} \|_2^2 
\quad \text{s.t} \ \|\mathbf{x}_n^{ij}\|_0 \leq T_n$ 
\If{n = N}
\State $\hat{\mathbf{Y}}_n \gets (\sum_{ij}\mathbf{R}^T_{ij}\mathbf{R}_{ij})^{-1}(\sum_{ij}\mathbf{R}^T_{ij}\hat{\mathbf{D}}_n\hat{\mathbf{x}}^{ij}_n) $
\Else
\State $\hat{\mathbf{Y}}_n \gets (\sum_{ij}\mathbf{R}^T_{ij}\mathbf{R}_{ij})^{-1}(\sum_{ij}\mathbf{R}^T_{ij}\hat{\mathbf{D}}_n\hat{\mathbf{x}}^{ij}_n)+ \text{upsample}(\hat{\mathbf{Y}}_{n+1},2)$
\EndIf
\EndFor
\State $\mathbf{Y}^{'} \gets \{\mathbf{Y}^{'}_N, \mathbf{Y}^{'}_{N-1}\cdots, \mathbf{Y}^{'}_0 \}$
\State $\forall {ij} \
  \hat{\mathbf{D}}_{\text{global}} \gets  \argmin\limits_{\mathbf{D}} \sum_{ij}\|\mathbf{R}_{ij}\mathbf{Y}^{'}-\mathbf{D}\mathbf{x}^{ij} \|_2^2 
 \quad \text{s.t} \ \|\mathbf{x}^{ij}\|_0 \leq T$ 
\State \textbf{Reconstruction:}
\State $\hat{\mathbf{Y}} \gets 0$
\For {$n = \{N,N-1,\cdots,0 \}$}
\State $\mathbf{Y}_n^{'} = \mathbf{Y}_n -\text{upsample}(\hat{\mathbf{Y}},2)$ 
\State  $\forall {ij} \
  \{ \hat{\mathbf{x}}^{ij}_n\} \gets  \argmin\limits_{\mathbf{x}_n^{ij}} \sum_{ij}\|\mathbf{R}_{ij}\mathbf{Y}_n^{'}-\hat{\mathbf{D}}_{global}\mathbf{x}_n^{ij} \|_2^2 
\quad \text{s.t} \ \|\mathbf{x}_n^{ij}\|_0 \leq T_n$ 
\State $\hat{\mathbf{Y}} \gets (\sum_{ij}\mathbf{R}^T_{ij}\mathbf{R}_{ij})^{-1}(\sum_{ij}\mathbf{R}^T_{ij}\hat{\mathbf{D}}_{global}\hat{\mathbf{x}}^{ij}_n)+ \text{upsample}(\hat{\mathbf{Y}},2)$
\EndFor
\State \textbf{return}
\end{algorithmic}
\end{algorithm}

\textbf{First-pass:}

We start at the coarsest layer $N$ in the cascade. After learning the layer dictionary and finding the sparse coefficients, we propagate consecutively the reconstructed images to the finer layers. In the coarsest layer we process the downsampled image, in the consecutive layers we encode and decode the residuals. In each layer, we use same size  ${b} \times {b}$ patches. A patch in the layer $n$ corresponds to a $({b} \times 2^n) \times ({b} \times 2^n)$ area in the original image. Algorithm~\ref{alg:psedo1} summarizes the first-pass. 

\textbf{Dictionary learning:} We learn a dictionary at the coarsest layer and use it to reconstruct the downsampled image. This layer's dictionary $\hat{\mathbf{D}}_N$ is produced by minimizing the objective function using the coarsest resolution image patches
\begin{align}\label{cost3}
\begin{split}
& \argmin_{\mathbf{D}_N,\mathbf{x}_N^{ij}} \sum_{ij}\|\mathbf{R}_{ij}\mathbf{Y}_N-\mathbf{D}_N\mathbf{x}_N^{ij} \|_2^2 + \lambda \|\mathbf{x}^{ij}_{N}\|_0 \\
\end{split}
\end{align}
where the operator $\mathbf{R}_{ij}$ is a binary matrix that extracts a square patch of size ${b} \times {b}$ at location $(i,j)$ in the image then arranges the patch in a column vector form. The parameter $\lambda$ balances the data fidelity term and the regularization term, and $\mathbf{x}^{ij}_{N}$ denotes the coefficients for the patch $(i,j)$ .

We initialize the dictionary $\mathbf{D}_N$ with a DCT basis by extracting several atoms from the DCT bases and applying Kronecker product. 
It is possible to choose any dictionary update methods such as KSVD~\cite{Aharon2006}, approximate KSVD~\cite{rubinstein2008efficient}, MOD~\cite{Engan1999}, and ODL~\cite{Mairal2009}. Both ODL and approximate KSVD can achieve the same PSNR with less coefficients. In order to reveal the strength of our method, we choose the original KSVD to update the dictionaries. Therefore, we do a sequence of rank-one approximations that update both the dictionary atoms and the coefficients. 

Iteratively, we first fix all coefficients $\hat{\mathbf{x}}_N^{ij}$ and select each dictionary atom one by one $\mathbf{d}_N^{l}$, $ l = \{1,2, \cdots,k\}$. For any atom $\mathbf{d}_N^{l}$, we extract the patches, which are composed by the atom $(i,j) \in \mathbf{d}_N^l$, to compute its residual. The corresponding coefficients are denoted as $\mathbf{x}^{ij}_N(l)$, which are the non-zero entries of the $l$-th row of coefficient matrix
\begin{align}
\mathbf{e}^{ij}_N(l) = \mathbf{R}_{ij}\mathbf{Y}_N- \hat{\mathbf{D}}_N\mathbf{x}_N^{ij} + \mathbf{d}^l_N\mathbf{x}^{ij}_N(l).
\end{align}
We arrange all $\mathbf{e}^{ij}_N(l)$ as the columns of the overall representation error matrix $\mathbf{E}_N^l$. Then, we update the atom $\hat{\mathbf{d}}^l_N$ and the $l$-th row $\hat{\mathbf{x}}_N(l)$ by
\begin{align}
\{ \hat{\mathbf{d}}^l_N,\hat{\mathbf{x}}_N(l)\} = \argmin_{\mathbf{d},\mathbf{x}}\|\mathbf{E}_N^l - \mathbf{d}\mathbf{x}\|_F^2.
\end{align}
Finally, we perform SVD decomposition on the error matrix, and update the $l$-th dictionary atom $\hat{\mathbf{d}}^l_N$ by the first column of $\mathbf{U}$, where $\mathbf{E}_N^l = \mathbf{U}\Sigma\mathbf{V}^T$. The coefficient vector $\hat{\mathbf{x}}_N(l)$ is the first column of matrix $\Sigma(1,1)\mathbf{V}$. In every iteration all dictionary atoms and coefficients are updated simultaneously.

\textbf{Sparse coding:} After getting the updated dictionary, sparse coding is done with the Orthogonal Matching Pursuit (OMP), a greedy algorithm that is computationally efficient~\cite{tropp2004greed}. The sparse coding stops when the number of coefficient reaches the upper limit $T_N$ or the reconstruction error becomes less than threshold

\begin{align}\label{sp}
\begin{split}
  \hat{\mathbf{x}}^{ij}_N  = & \argmin_{\mathbf{x}_N^{ij}} \sum_{ij}\|\mathbf{R}_{ij}\mathbf{Y}_N^{'}-\hat{\mathbf{D}}_n\mathbf{x}_n^{ij} \|_2^2 \quad  \text{s.t.} \quad \|\mathbf{x}^{ij}_{n}\|_0 \leq T_N. \\
\end{split}
\end{align}
Putting the updated coefficient matrix $\hat{\mathbf{x}}^{ij}_N$ back into Dictionary learning to update the dictionary and coefficient until reaching the iteration times.

\textbf{Residuals:} In each layer, we use at most $T_n$ active coefficients for each patch to reconstruct the image and then compute the residual. The number of coefficients governs how strong the residual to emerge. Larger values of $T_n$ generates a more accurate reconstructed image. Thus, the total energy of residuals will diminish. Smaller values of $T_n$ cause the residual to increase, not only due to sparse coding but also resampling across layers. Since the dictionary is designed to represent a wide spectrum of patterns to keep the encodings as sparse as possible, $T_n$ should be small. The reconstructed image is a weighted average of the patches that contain the same pixel
\begin{align}
\begin{split}
\hat{\mathbf{Y}}_N = (\sum_{ij}\mathbf{R}^T_{ij}\mathbf{R}_{ij})^{-1}(\sum_{ij}\mathbf{R}^T_{ij}\hat{\mathbf{D}}_N\hat{\mathbf{x}}^{ij}_N). 
\end{split}
\end{align}
After decoding based on the dictionary $\hat{\mathbf{D}}_{N}$, we obtain the residual image $\mathbf{Y}^{'}_{N-1}$ by subtracting the upsampled reconstruction $\mathbf{U}(\hat{\mathbf{Y}}_N)$  from the next layer image $\mathbf{Y}_{N-1}$, e.g. $ \mathbf{Y}_{N-1}^{'}  =  \mathbf{Y}_{N-1}  -  \mathbf{U}(\hat{\mathbf{Y}}_{N})$. Here,
$\mathbf{U}(\cdot)$ denotes the bicubic upsampling operator. As the procedure of dictionary learning ans sparse coding in the $N$-th layer, we reconstruct residual $\hat{\mathbf{Y}}^{'}_{N-1}$ by training a residual dictionary $\mathbf{D}_{N-1}$ from the residual image itself. We keep encoding and decoding on residuals up to the finest layer. The cascade residual dictionary learning and reconstruction can be expressed as follows:
\begin{align}\label{recon}
\begin{split}
 \{ \hat{\mathbf{x}}^{ij}_n, \hat{\mathbf{D}}_n\} = & \argmin_{\mathbf{x}_n^{ij}, \mathbf{D}_n} \sum_{ij}\|\mathbf{R}_{ij}\mathbf{Y}_n^{'}-\mathbf{D}_n\mathbf{x}_n^{ij} \|_2^2 \quad  \text{s.t.} \quad \|\mathbf{x}^{ij}_{n}\|_0 \leq T_n, \\
\end{split}
\end{align}
where residual image is
\begin{align}
\mathbf{Y}_n^{'} = \left\{
\begin{array}{lll}
 \mathbf{Y}_n - \mathbf{U}(\hat{\mathbf{Y}}_{n+1}), & \qquad  & 0 \leq n < N \\
 \mathbf{Y}_N, & \qquad  &  n = N, \\
\end{array}
\right.
\end{align}
and the reconstructed residual is 
\begin{align} \label{reconstruction}
\hat{\mathbf{Y}}_n = 
\left\{
\begin{array}{lll}
 (\sum_{ij}\mathbf{R}^T_{ij}\mathbf{R}_{ij})^{-1}(\sum_{ij}\mathbf{R}^T_{ij}\hat{\mathbf{D}}_n\hat{\mathbf{x}}_{n}^{ij}) +  \mathbf{U}(\hat{\mathbf{Y}}_{n+1}), & \quad   &  0 \leq n < N \\
 (\sum_{ij}\mathbf{R}^T_{ij}\mathbf{R}_{ij})^{-1}(\sum_{ij}\mathbf{R}^T_{ij}\hat{\mathbf{D}}_n\hat{\mathbf{x}}_{n}^{ij}), & \qquad  &  n = N. \\
\end{array}
\right.
\end{align}


The more coefficients used, which reduce the error caused by sparse representation. Since we are pursuing sparser representation, less number of coefficient would be better.





\textbf{Second-pass:}
\begin{figure}[t]\label{secpassfig}
 \centering
 \includegraphics[width=3in]{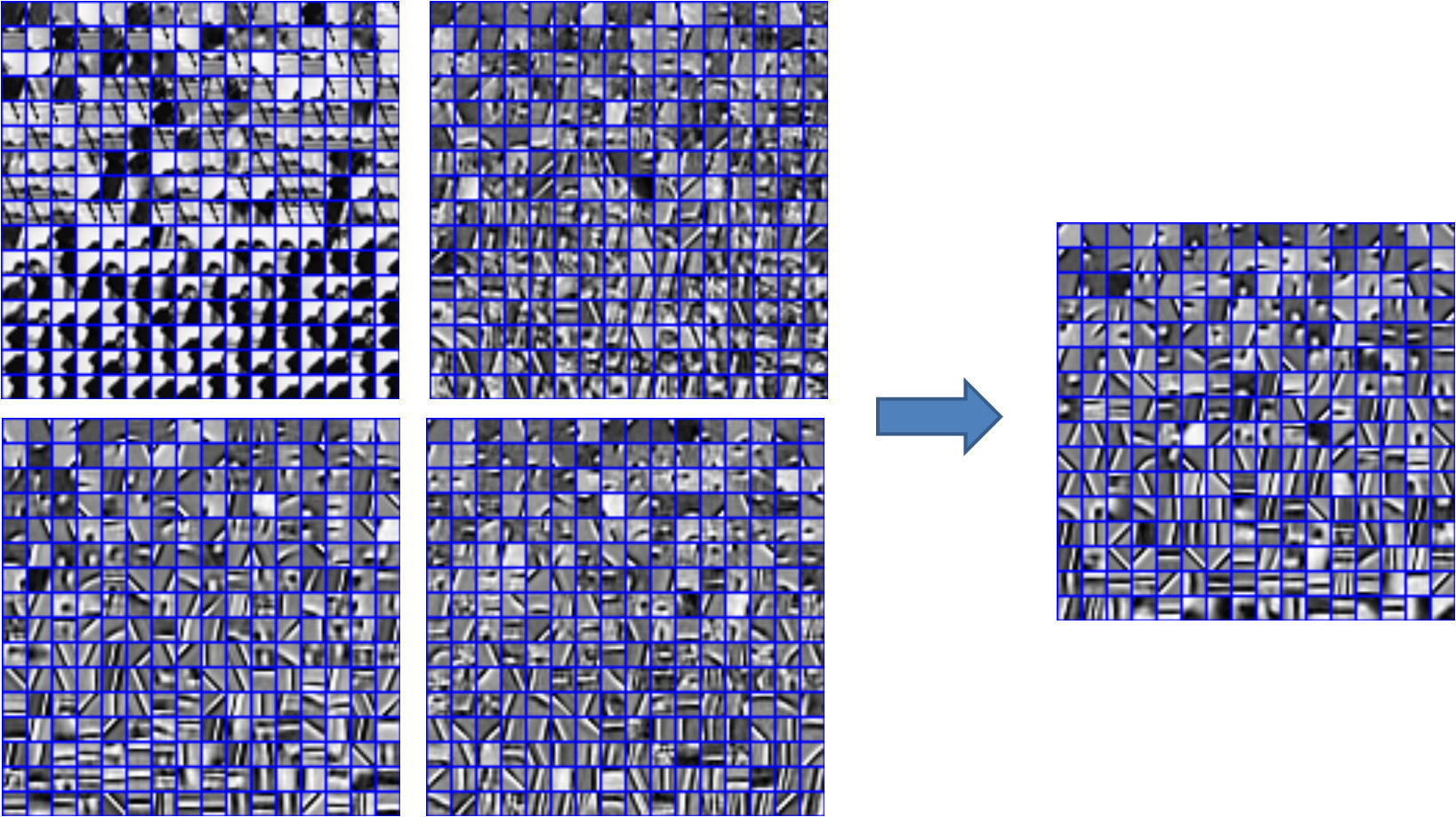}
 \caption{Left: Four dictionaries of the different levels learned in the first pass (clockwise from the upper-left: the coarsest level, the second, the third, and the finest level). Right: The unifying dictionary learned in the second pass.}
\end{figure} 

In each layer the more atoms we use, the better quality can be achieved. However, this would not be the best use of the limited number of atoms. For instance, image patches from the coarsest layer are limited both in quantity and variety. The residual images are relatively sparse which imply they do not require many dictionary atoms. However, it is not straightforward to determine the optimal number of atoms for each dictionary since finer level residuals depend on coarser ones. Rather than keeping all dictionaries, we train a global dictionary $\mathbf{D}$ using patches from $\mathbf{Y}^{'} = \{\mathbf{Y}_N,\mathbf{Y}^{'}_{N-1}, \cdots, \mathbf{Y}^{'}_{0} \}$. As illustrated in Fig.~\ref{secpassfig}, the dictionaries learned from $\mathbf{Y}^{'}$ in the first pass are redundant. The overall dictionary is less repetitive and more general to reconstruct all four layers. This allows us to select most useful atoms automatically without making sub-optimal layer-wise decisions. Notice that, in this procedure the number of coefficient can be arbitrarily chosen depending on the target quality of each layer.

\section{Experimental Analysis} \label{experiments}

To demonstrate the flexibility of our method, we evaluate its performance on three different and popular image processing tasks: image coding, image denoising, and image inpainting. Our method is shown to generate the best image inpainting results and provide the most compact set of coding coefficients.

\subsection{Image Coding}

We compare our method with five state-of-the-art dictionary learning algorithms including both single and multi-scale methods: approximate KSVD (a-KSVD)~\cite{rubinstein2008efficient}, ODL~\cite{Mairal2009}, KSVD~\cite{Aharon2006} Multi-scale KSVD~\cite{mairal2008learning}, Multi-scale KSVD using wavelets (Multi-wavelets)~\cite{Ophir2011}. 

For objectiveness, we use the same number of dictionary atoms for our and all other methods. Notice that, a larger dictionary would generate a sparser representations. We employ 4-times over-complete dictionaries, i.e. $\mathbf{D} \in \mathbb{R}^{64 \times 256}$ except for the Multi-wavelets where the dictionary in each sub-band has as many atoms as our dictionary (in favor of Multi-wavelets).


\begin{figure}[t]
 \centering
  \subfigure[animals]{
    \includegraphics[width=1.5in]{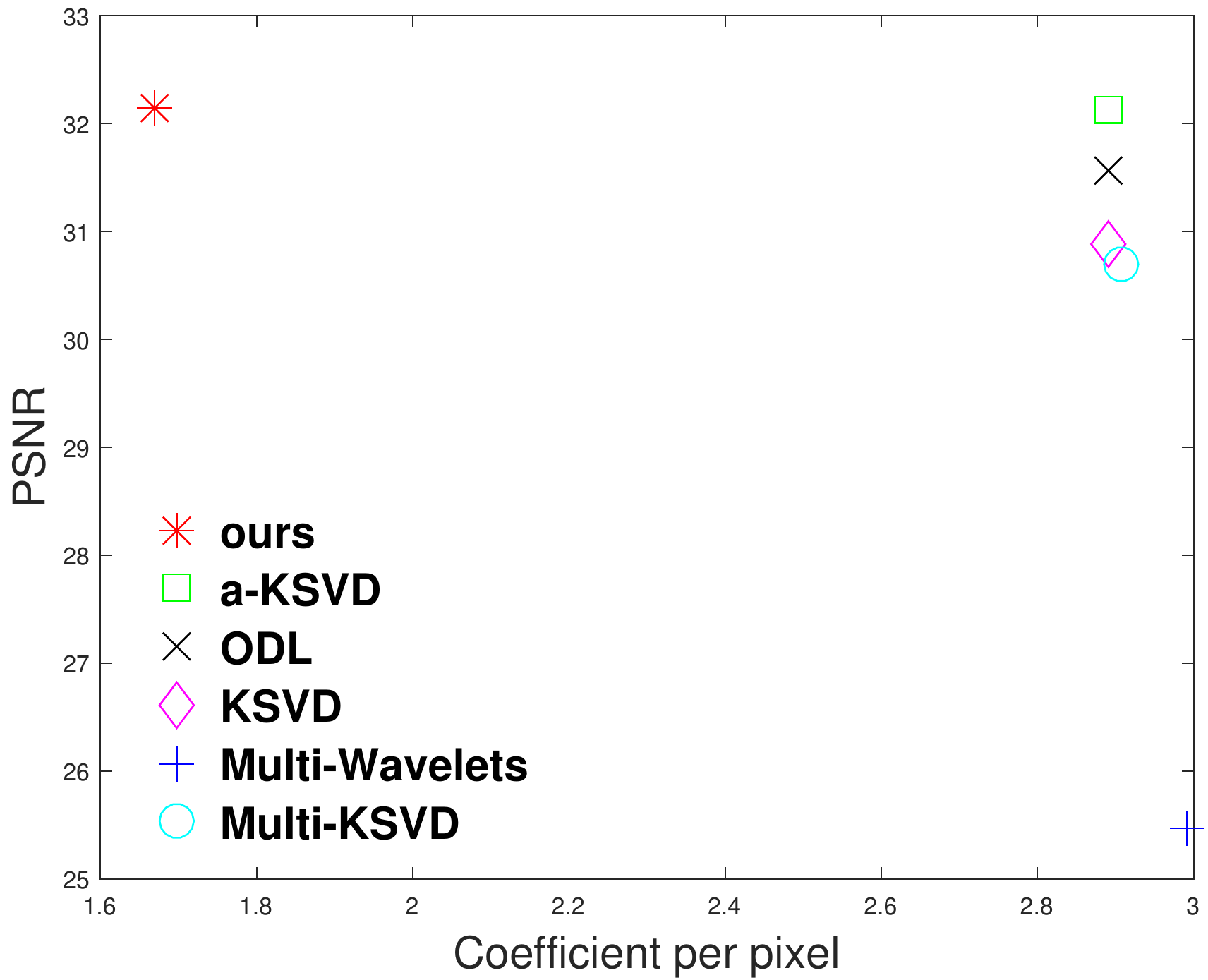}}
  \hspace{1cm}
  \subfigure[landscape]{
    \includegraphics[width=1.5in]{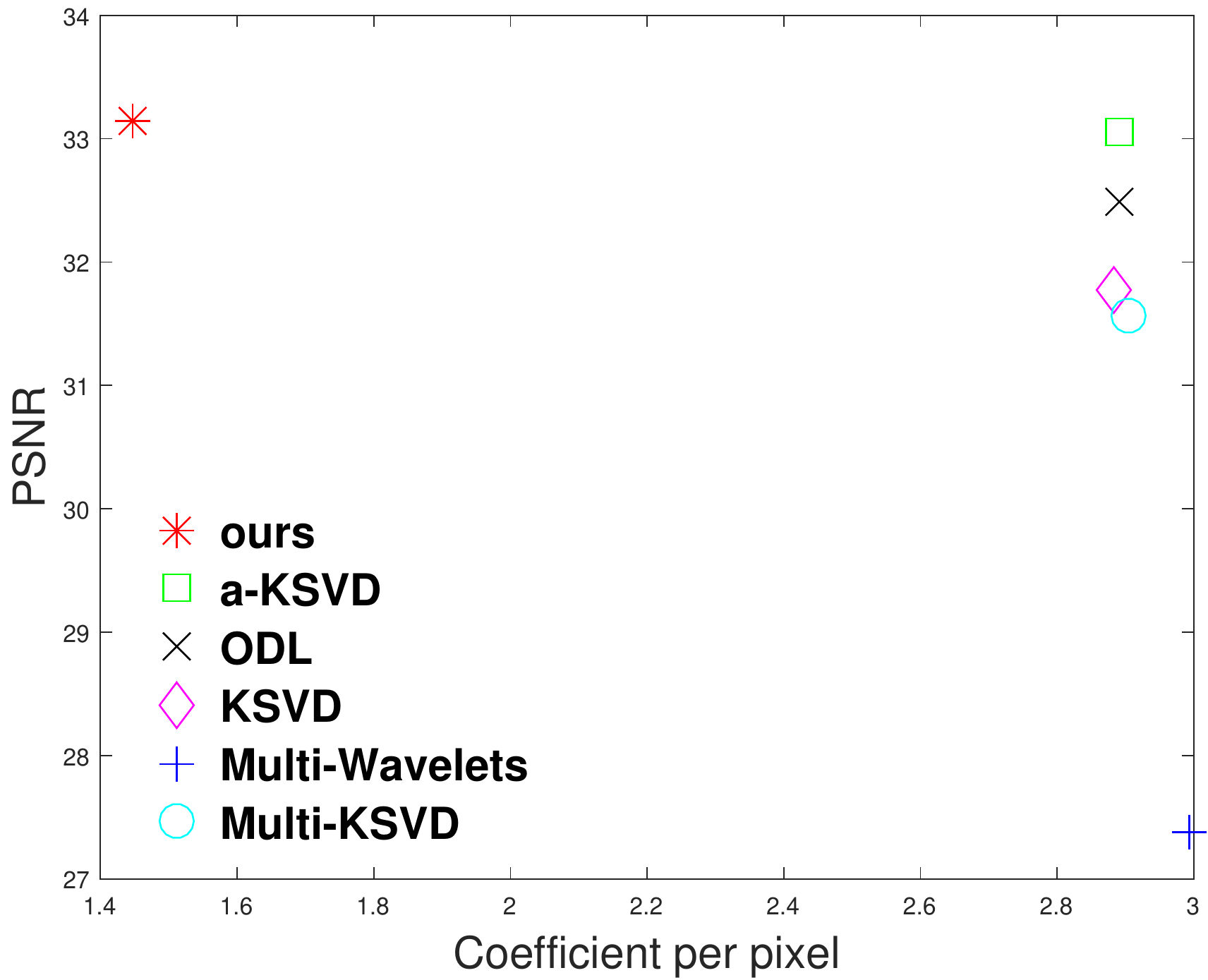}}
    \hspace{1cm}
  \subfigure[texture]{
    \includegraphics[width=1.5in]{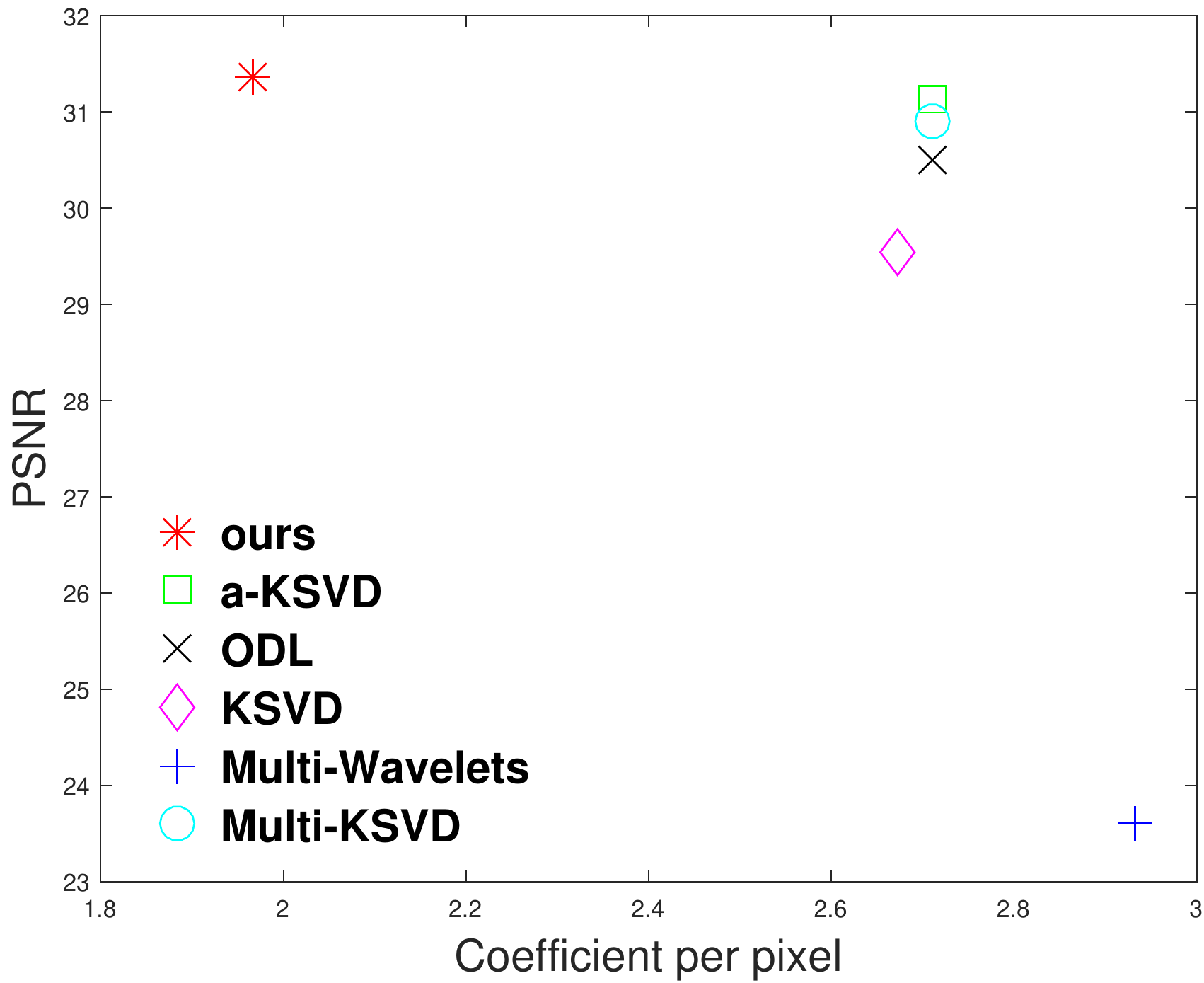}}
    \subfigure[face]{
    \includegraphics[width=1.5in]{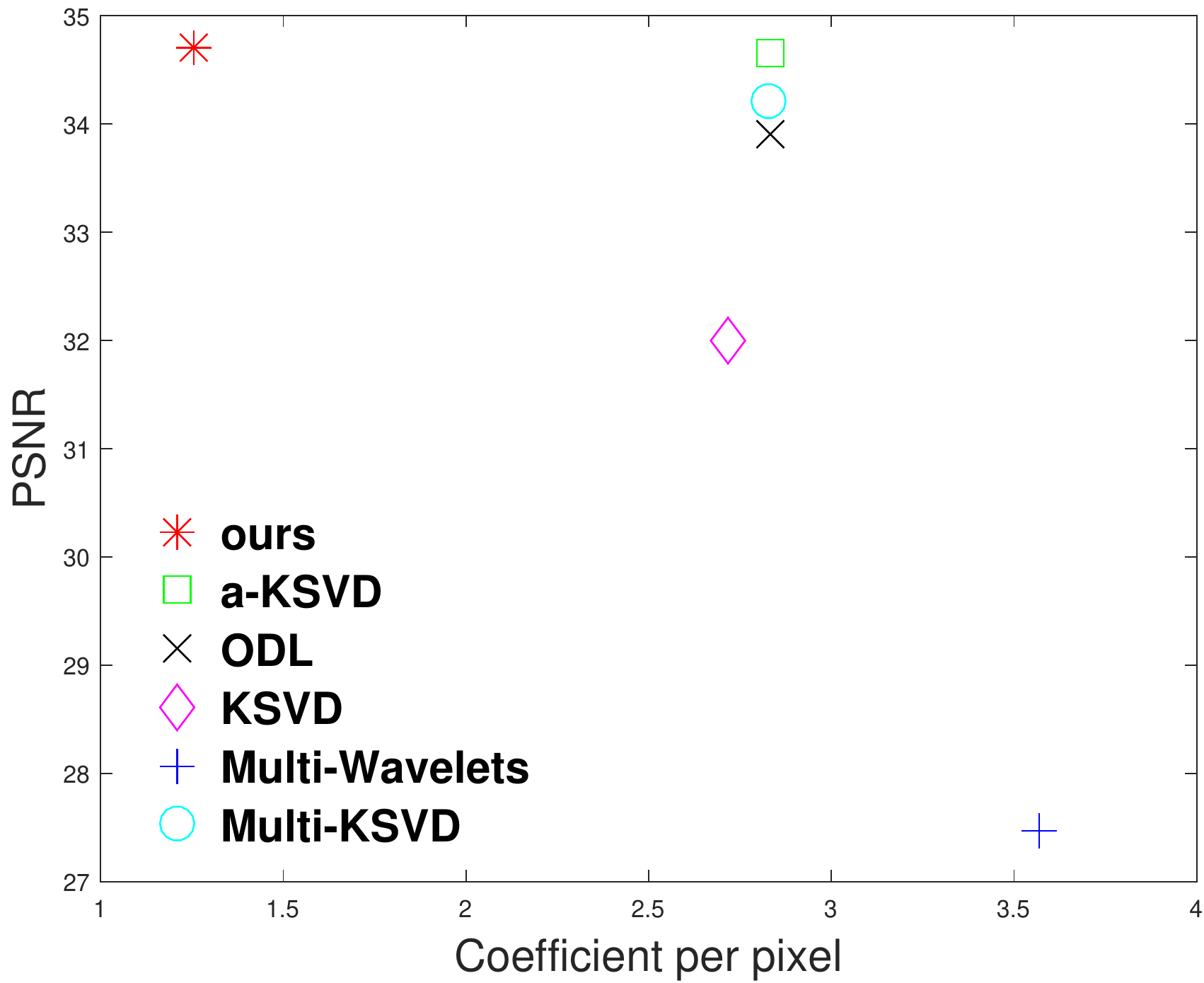}}
    \subfigure[fingerprint]{
    \includegraphics[width=1.5in]{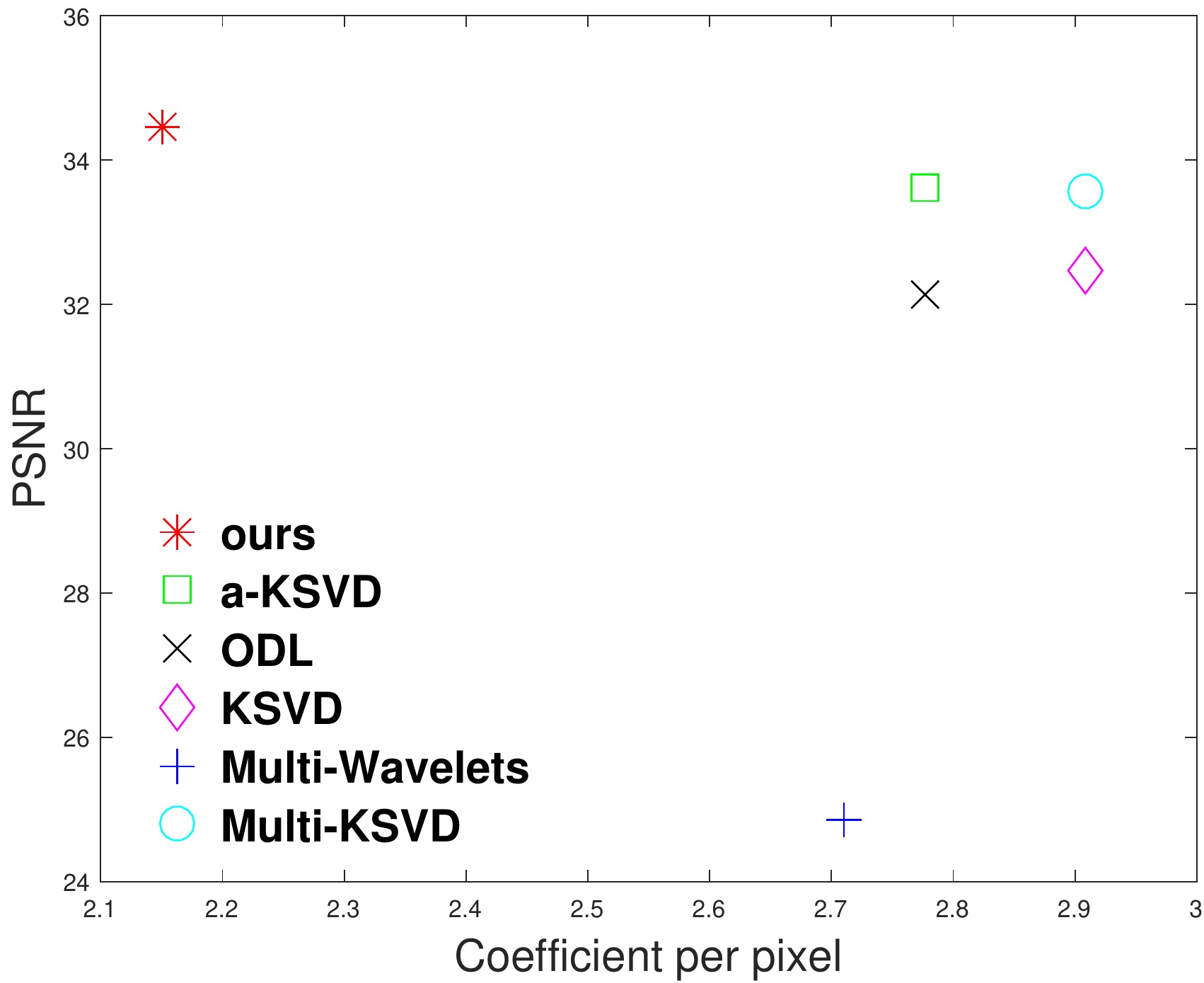}}
\caption{Reconstruction results on different 5 different image datasets. The horizontal axis represents the number of coefficient per pixel and the vertical axis is the quality in terms of PSNR (dB).}
  \label{fig5} 
\end{figure} 

For a comprehensive evaluation, we build five different image datasets, each contains 50 samples of a specific class of images: animals, landscape, texture, face, and fingerprint. 

Figure~\ref{fig5} depicts the number of coefficients per pixel vs. PSNR as the function of number of coefficient per each pixel. Each point is the average score for the corresponding method. As seen, our method is the best performing algorithm among the state-of-the-art. In all five image datasets, it achieves higher PSNR scores with significantly much less number of coefficients. In these experiments, the patches are extracted by 1-pixel overlapping in all images. We use $8 \times 8 $ blocks on each layer, and the cascade comprises 4 layers. Since the blocks in every layer have the same size, the lower resolution blocks efficiently represent larger receptive fields when they are upsampling onto a higher resolution. 

From another perspective, when decoding on the coarsest resolution, our method first employs $8 \times 8$ blocks, which corresponds to $8*2^{n-1} \times 8*2^{n-1}$ region on the finest resolution using the same dictionary atoms. Since there is a single global dictionary after the second pass, all layers share the same atoms. This resembles the quadtree structure, however, our method is not limited by the size of the dictionary (dimension of patches - atoms - and number of atoms) and it is as fast as single-scale dictionary learning and sparse coding. For Multiscale KSVD, the maximum dimension of dictionary atom can be 8 and only 2 scales can be performed. Thus, we extracted 128 atoms at each scale. 

\begin{figure}[t]
 \centering
  \subfigure[a-KSVD:28.68 db PSNR]{
    \includegraphics[width=2.2in]{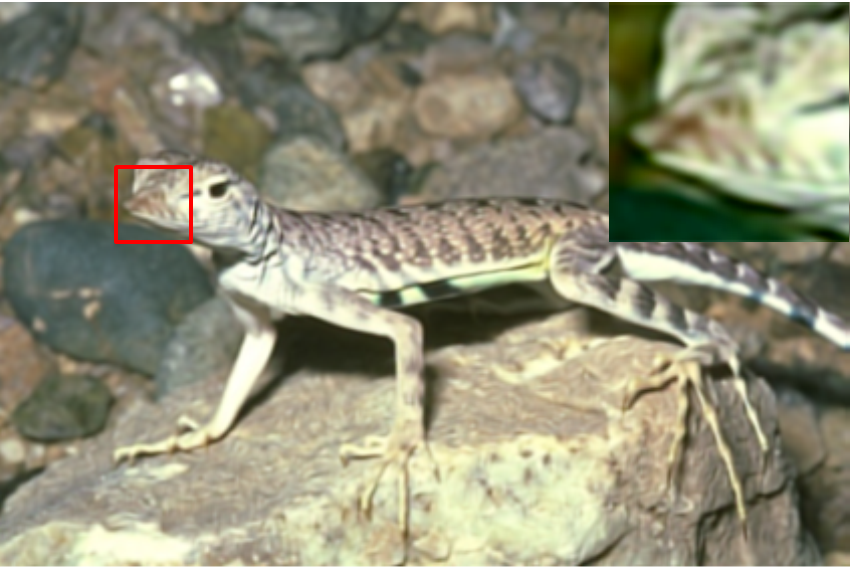}}
    \subfigure[Our method: 32.62 db PSNR]{
    \includegraphics[width=2.2in]{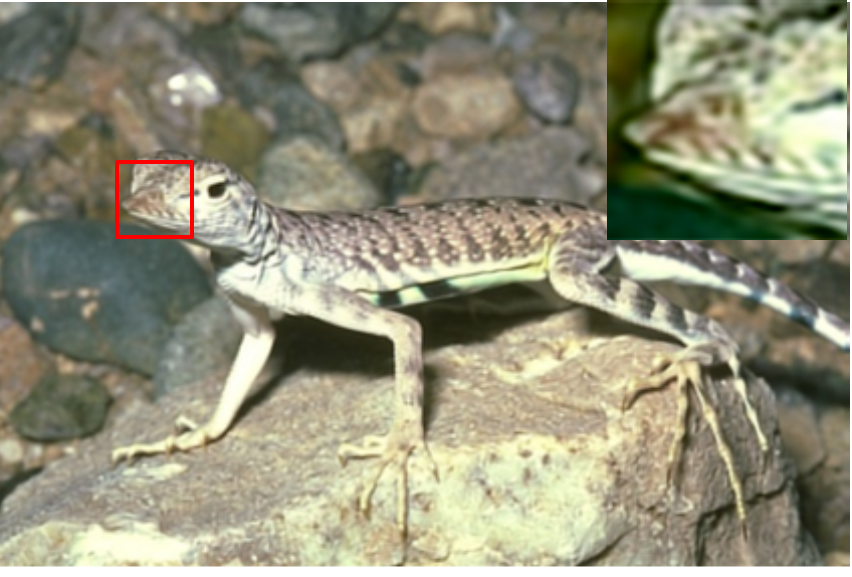}}    
\caption{Image coding results the comparison between a-KSVD and our method. Our method uses 1309035 coefficients and achieves 32.62 db PSNR score while a-KSVD uses 1332286 coefficients to get 28.65 dB PSNR. our method is almost \textbf{4 dB} better. Enlarged red regions are shown on the top-right corner of each image. As visible, our method produces more accurate reconstructions. }  
  \label{fig6}
\end{figure}

Compared with other algorithms, our method can save an outstanding $55.6\%$, $42.23\%$ and $49.95\%$ coefficients for the face, animals, and landscape datasets, respectively. For the image classes where spatial texture is dominant, our method is also superior by decreasing the number of coefficient by $27.74\%$ and $22.38 \%$ for the texture and fingerprint datasets. Sample image coding results for qualitative assessment are given in Fig.~\ref{fig6}. As shown, a-KSVD image coding is inferior to our even though a-KSVD uses more coefficients.


\subsection{Image denoising}

We also analyze the image denoising performance of our method. We make comparison with five dictionary learning algorithms. We note that the state-of-the-art is collaborative and non-local techniques, such as BM3D~\cite{dabov2007image}, LSSC~\cite{mairal2009non}, yet we do not engineer a collaborative scheme. Our goal here is to understand how our method compares to other dictionary learning methods. 

\begin{figure}[!ht]
\centering
\includegraphics[width=1.0\textwidth]{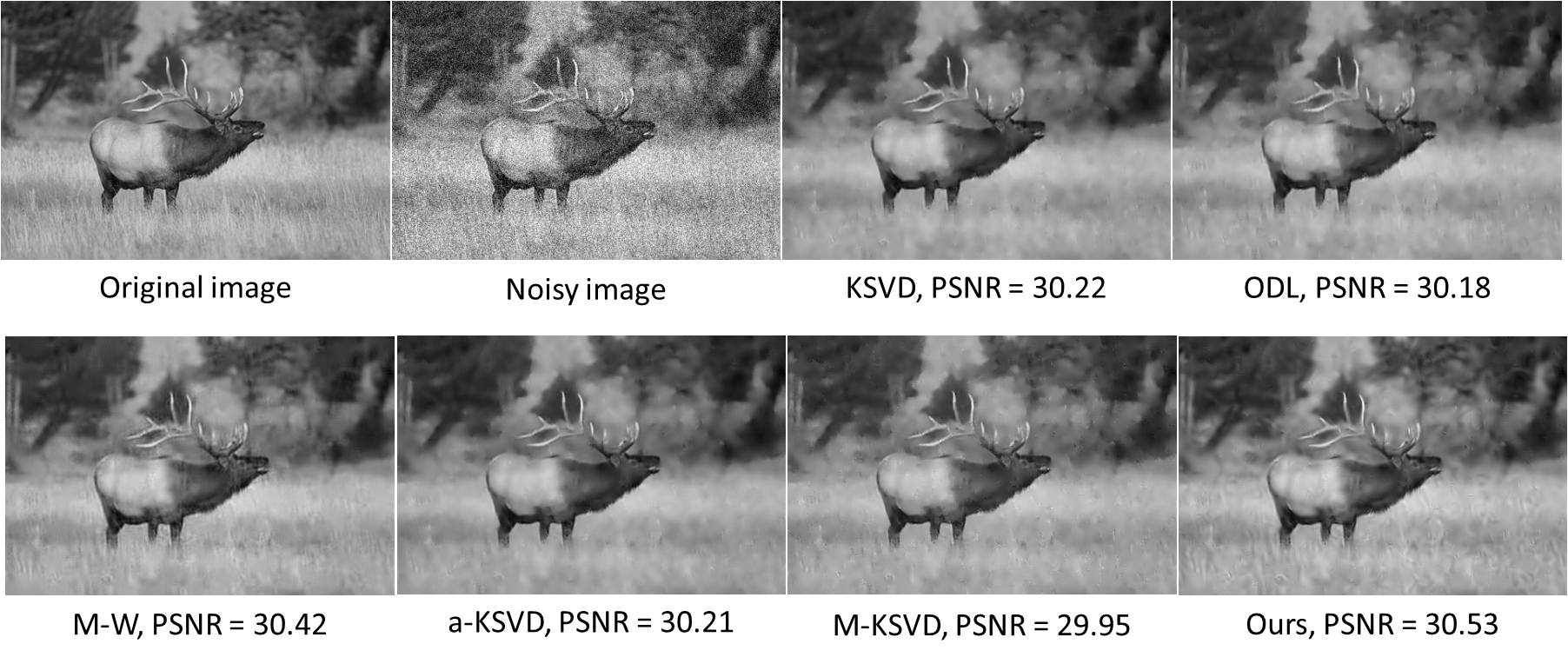}
\includegraphics[width=1.0\textwidth]{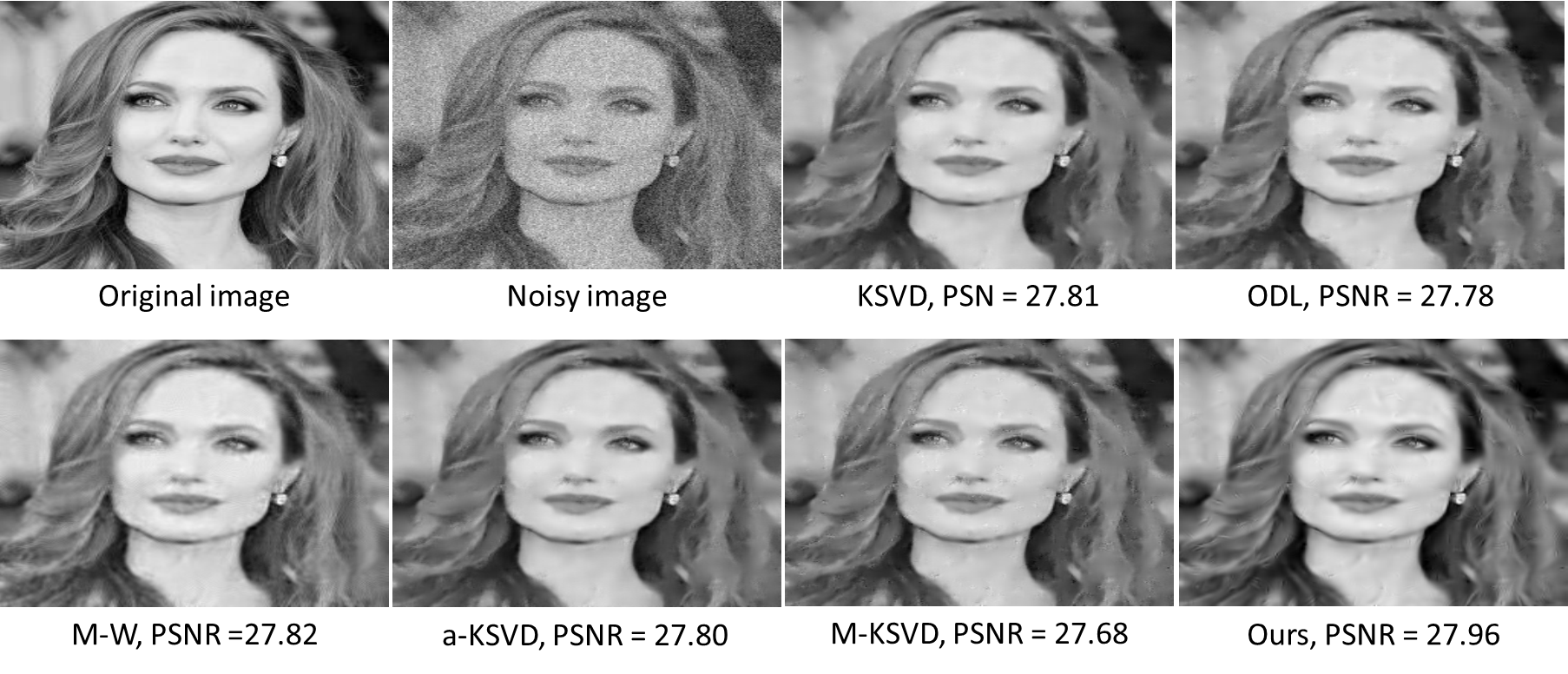}
\vspace{-3mm}
\caption{Denoised face images. Additive zero-mean Gaussian noise with $\sigma = 30$.}\label{denoise2}
\end{figure}

We minimize the cost function in Eqn.~\eqref{eq:experiments:denoise} for denoising. We use the difference between the downsampled input image and aggregated reconstructions at each layer to terminate the OMP.
\begin{align}
\begin{split}
& \argmin_{\mathbf{x}_i} \sum_{ij}\|\mathbf{x}^{ij}_{n}\|_0\\
& \text{s.t.} \| \mathbf{R}_{ij}\mathbf{Y}_n-\mathbf{D}_n\mathbf{x}_n^{ij} + \mathbf{R}_{ij}\mathbf{U}(\hat{\mathbf{Y}}_{n+1}) \|_2^2 \leq  C{\sigma}
\end{split}
\label{eq:experiments:denoise}
\end{align}
Above, the reconstructed residual $\hat{\mathbf{Y}}_{n+1}$ is defined as in Eqn.~$\eqref{reconstruction}$, and $\sigma$ is chosen according to the variance of the noise. As before, we choose the 4-layer cascade and $8 \times 8$ patch size. The parameters of KSVD and multi-scale wavelets are set as recommenced by original authors. We fixed all parameters for all test images. As shown in Fig.~\ref{denoise2}, our method achieves higher PSNR scores than the state-of-the-art. In addition, it can render finer details more accurately.



\subsection{Image Inpainting}

Image inpainting is often used for restoration of damaged photographs and removal of specific artifacts such as missing pixels. Previous dictionary learning based algorithms work when the missing area is small and smaller than the dimension of dictionary atoms. 

As demonstrated in Fig.~\ref{fig:subfig:inpainting2} our method can restore the missing image regions that are remarkably much larger than the dimension of dictionary atoms, outperforming the state-of-the-art methods. By reconstructing the image starting at the coarsest layer, we can fix completely missing regions. The larger the missing area, the smoother the restored image becomes. In comparison, single-scale based methods fail completely. 

Given the mask $\mathbf{M}$ of missing pixels,  our formulation in each layer is
\begin{align}
\begin{split}
\hat{\mathbf{x}}_n^{ij} = \argmin_{\mathbf{x}_n} \sum_{ij}\|\mathbf{R}_{ij}\mathbf{M} \otimes (\mathbf{R}_{ij}\mathbf{Y}_n^{'}-\mathbf{D}_n\mathbf{x}_n )\|_2^2 
\\
\text{subject to} \quad \|\mathbf{x}^{ij}_{n}\|_0 \leq T_n 
\end{split}
\end{align}
where we denote $\otimes$ as the element-wise multiplication between two vectors. 

\begin{figure}[!ht]
  \centering
  \subfigure[original image]{
    \includegraphics[width=2.3in]{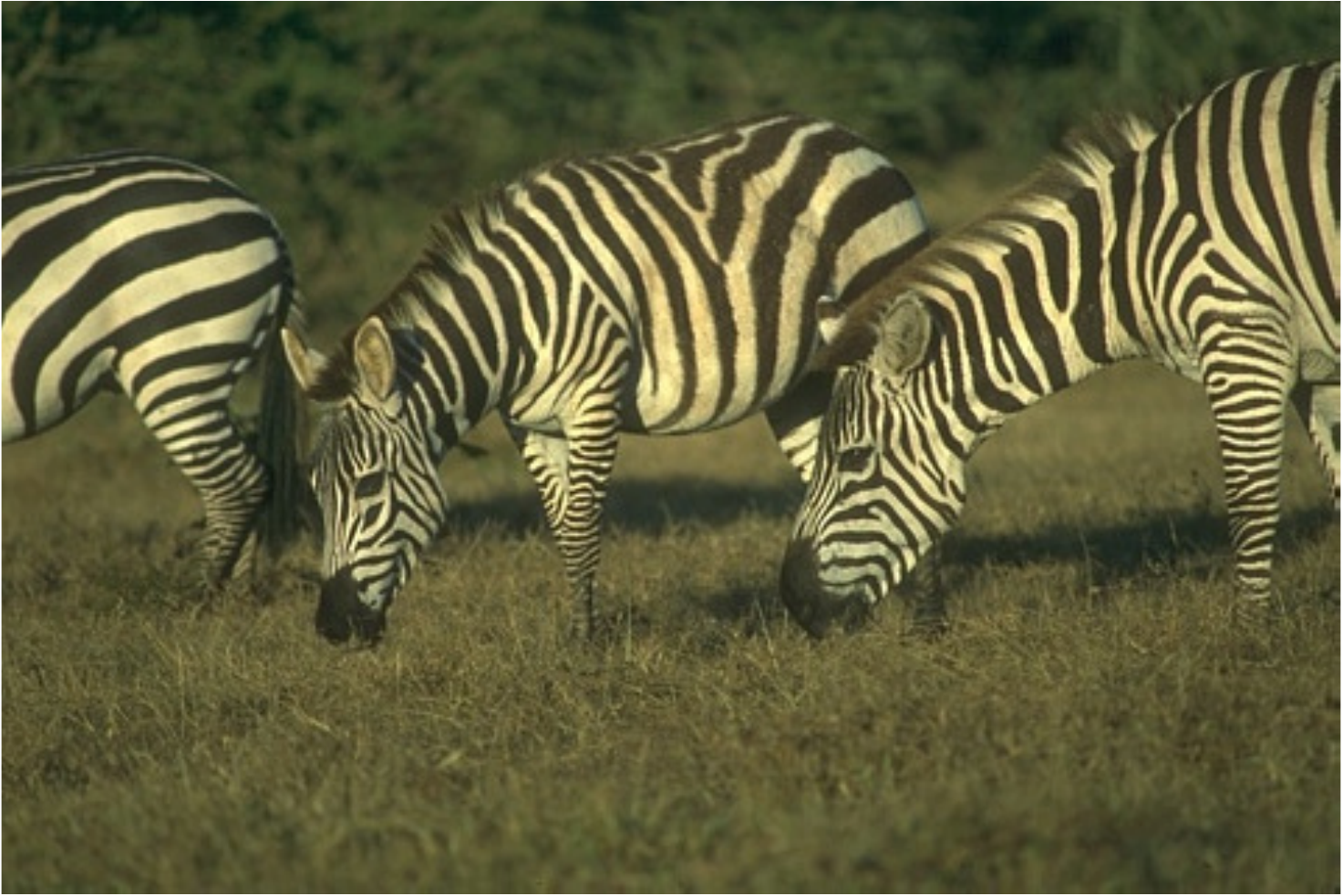}}
    \subfigure[corrupted image]{
    \includegraphics[width=2.3in]{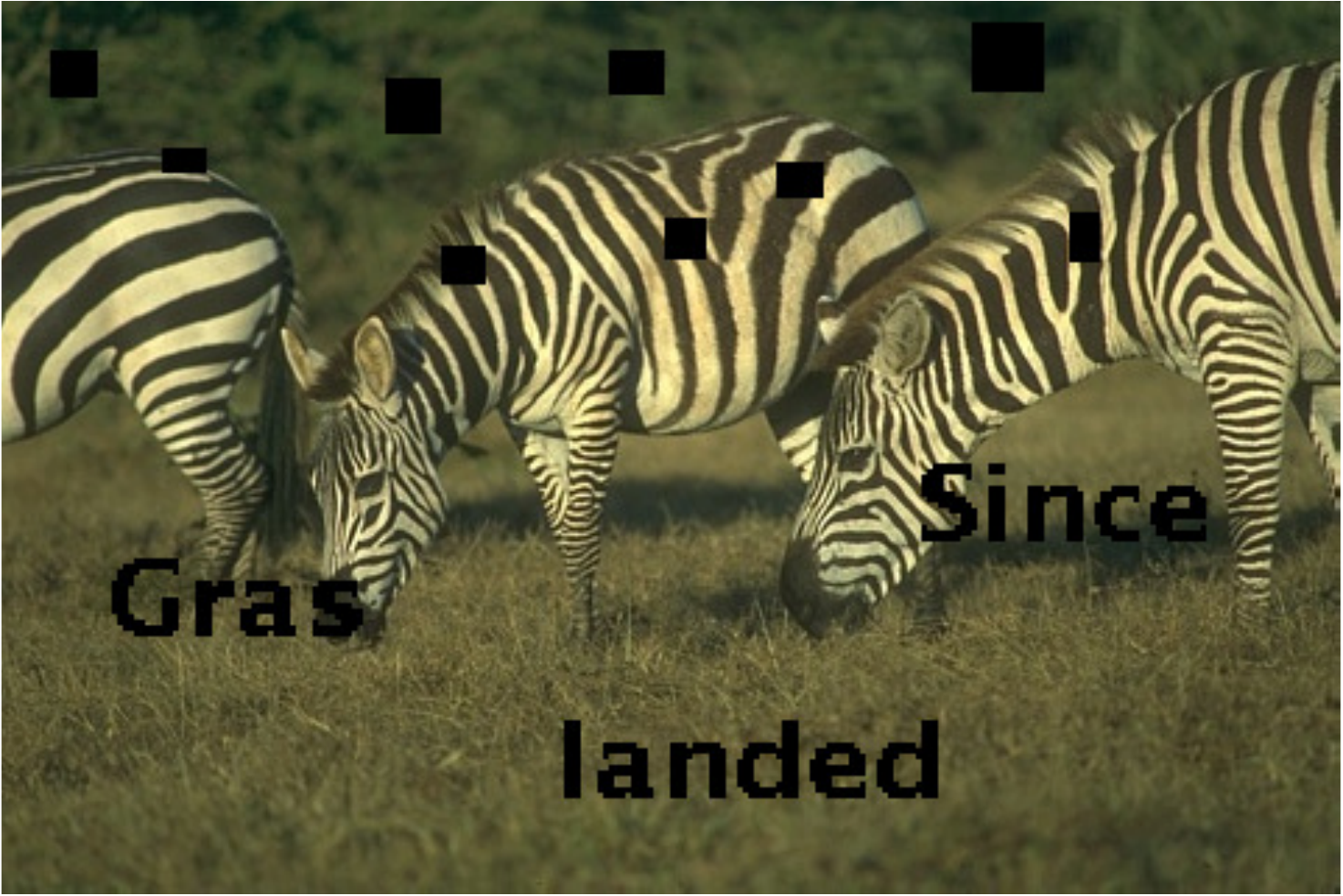}}
  \subfigure[KSVD]{
    \includegraphics[width=2.3in]{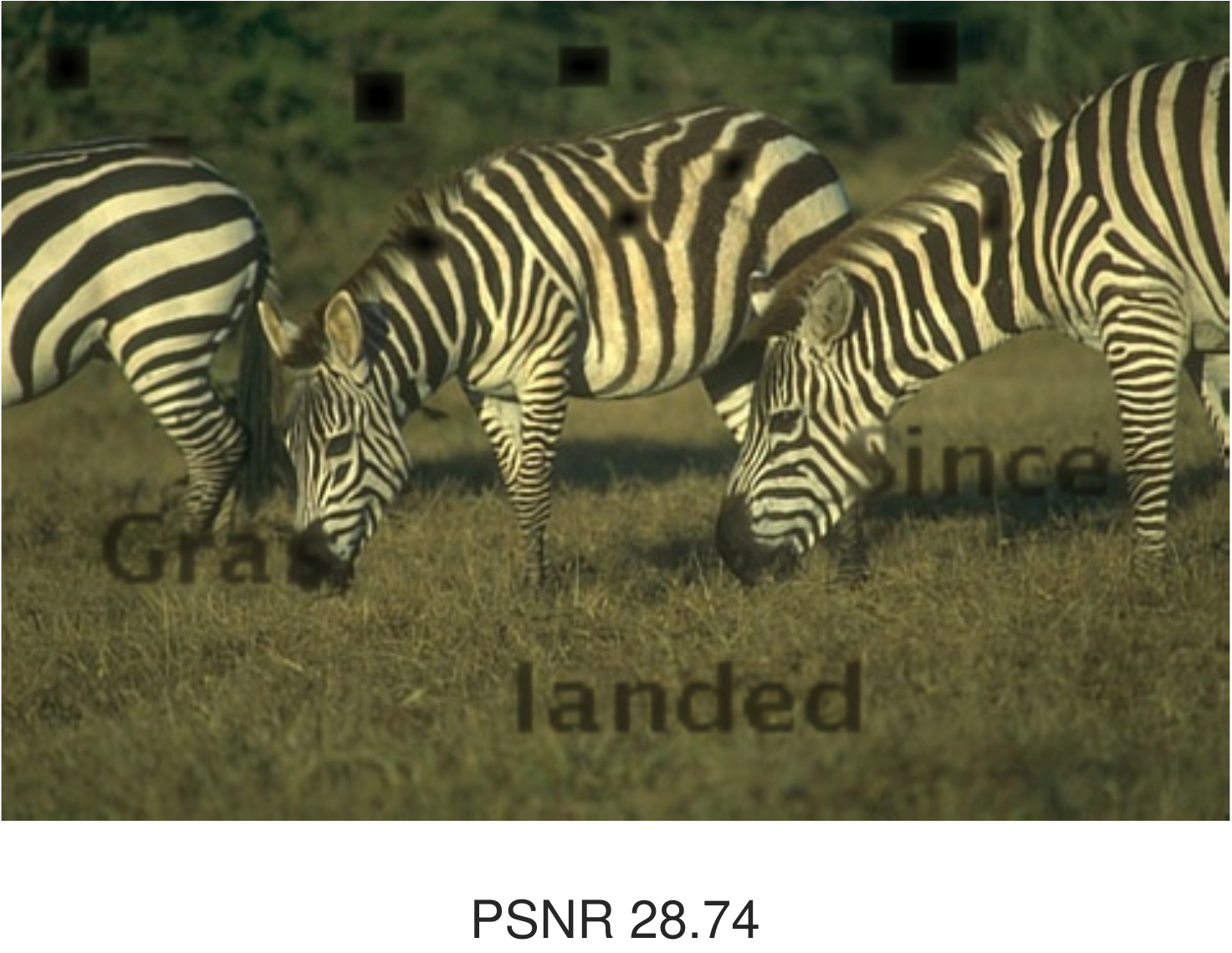}}
  \subfigure[Ours]{
    \includegraphics[width=2.3in]{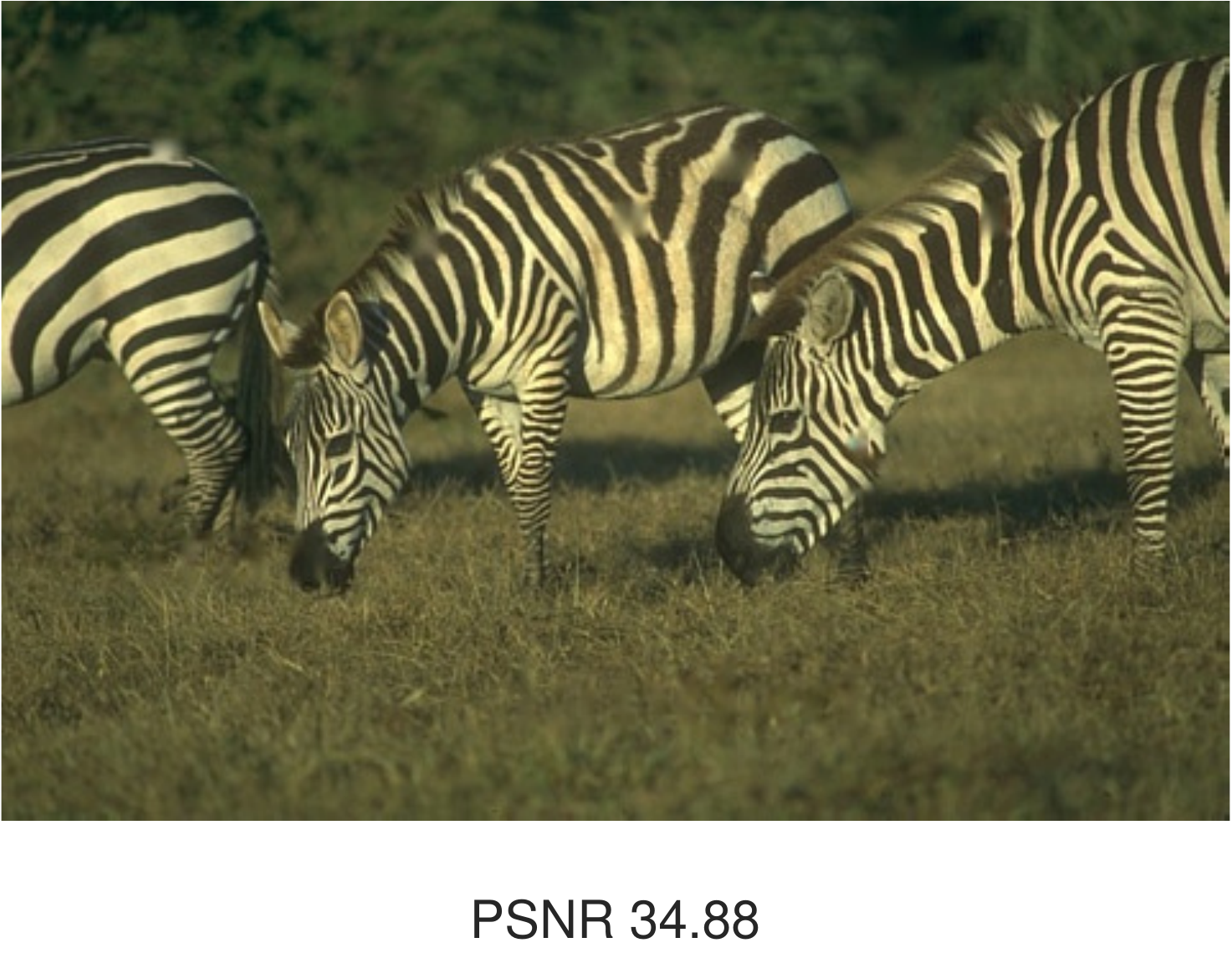}}
  \caption{The original image is corrupted with large artifacts. The sizes of the artifacts range from 8 to 32 pixels. Our method efficiently removes the artifacts. }
  \label{fig:subfig:inpainting1} 
\end{figure}

Figure~\ref{fig:subfig:inpainting1} shows that our algorithm can fix big holes and gaps but the KSVD can not. In this experiments we only compare with KSVD algorithm, because multiscale KSVD simply increases the dimension of atoms, which leads proportionally more atoms to form an overcomplete dictionary. At the same time, multiscale KSVD still fails to handle holes larger than the dimension of atoms.

\section{Conclusion}
\label{conclude}
\vspace{-2mm}
We presented a dictionary learning and sparse coding method on cascaded residuals. Our cascade allows capturing both local and global information. Its coarse-to-fine structure prevent from reconstructing the regions that can be well represented by the coarser layers. Our sparse coding can be used to progressively improve the quality of the decoded image. 

Our method provides significant improvement over the state-of-the-art solutions in terms of the quality of reconstructed image, reduction in the number of coefficients, and computational complexity. It generates much higher quality images using less number of coefficients. It produces superior results on image inpainting, in particular, in handling of very large ratios of missing pixels and large gaps. 

\section*{Acknowledgment}
\vspace{-2mm}
This work was supported by the Australian Research Council`s Discovery Projects funding scheme (project DP150104645).



\bibliographystyle{splncs}
\bibliography{accv2016submission.bib}

\end{document}